\documentclass{article}
\usepackage{bbm}
\usepackage{graphicx}
\usepackage{amsmath}
\usepackage{amssymb}
\usepackage{hyperref}
\usepackage{makecell}
\usepackage{multicol}
\usepackage{natbib}
\usepackage{acl}
\usepackage{xcolor}

\usepackage{caption}
\usepackage{subcaption}

\newcommand{\para}[1]{\vspace{3pt}\noindent\textbf{#1}}

\author{Halley Young \\
  Microsoft Research \\
  \texttt{halleyyoung@microsoft.com} \\\And
  Yimeng Zeng \\
  University of Pennsylvania \\
  \texttt{yimengz@seas.upenn.edu} \\ 
  \AND
  Jacob Gardner \\
  University of Pennsylvania \\
  \texttt{jacobrg@seas.upenn.edu} \\ \And
  Osbert Bastani \\
  University of Pennsylvania \\
  \texttt{obastani@seas.upenn.edu}
}

\title{Improving Structural Diversity of Blackbox LLMs via Chain-of-Specification Prompting}

\begin{document}

\maketitle
\begin{abstract}
The capability to generate diverse text is a key challenge facing large language models (LLMs). Thus far, diversity has been studied via metrics such as $n$-gram diversity or diversity of BERT embeddings. However, for these kinds of diversity, the user has little control over the dimensions along which diversity is considered. For example, in the poetry domain, one might desire diversity in terms of rhyme and meter, whereas in the code domain, one might desire diversity in terms of the kinds of expressions used to solve a problem. We propose a diversity metric called \emph{structural diversity}, where the user provides a mapping from generated text to features capturing the kinds of diversity that they care about. In addition, we propose a novel strategy called \emph{chain-of-specification (CoS) prompting} for improving diversity by first having the LLM generate a specification encoding one instance of structural features, and then prompting the LLM to generate text that satisfies these features; notably, our strategy works with blackbox LLMs. In our experiments, we show that for structural diversity in the poetry and code domains, CoS significantly improves diversity compared to several baselines.
\end{abstract}

\section{Introduction}

Recent advances in large language models (LLMs), such as ChatGPT \cite{chatgpt}, have led to significant improvements in the quality and coherence of machine-generated text. However, the diversity of the generated text remains limited, particularly in terms of capturing high-level semantic properties and stylistic variations. As a consequence, there has been a great deal of interest in techniques for improving the diversity of LLMs.

Much of the existing work on diversity has focused on metrics based on $n$-grams or semantic representations such as BERT embeddings. However, in many applications, users may desire diversity along specific dimensions. For instance, users might want generated poems to be diverse in terms of the structure and content of the poem, such as imagery and language, rhyming scheme, meter, etc. Alternatively, in code generation, users may want to generate code in using a range of different paradigms (e.g., for Python, list comprehension vs. loop vs. recursion) so they can choose the fastest.

To account for these forms of diversity, we assume the user has provided a feature mapping $\phi:\mathcal{X}\to\mathcal{S}$ that maps text $x\in\mathcal{X}$ to a feature vector $\phi(x)\in\mathcal{S}=\{0,1\}^d$. Then, we can measure diversity in terms of entropy of the generated text in feature space. In particular, given a number of random generations $\{x_1,...,x_k\}$, we can use the empirical entropy of the distribution $\{\phi(x_1),...,\phi(x_k)\}$. We focus on mappings $\phi$ that encode structural properties of text, such as the example structures of poems and programs given above; thus, we refer to this notion of diversity as \emph{structural diversity}.

\begin{figure}
\includegraphics[width=0.48\textwidth]{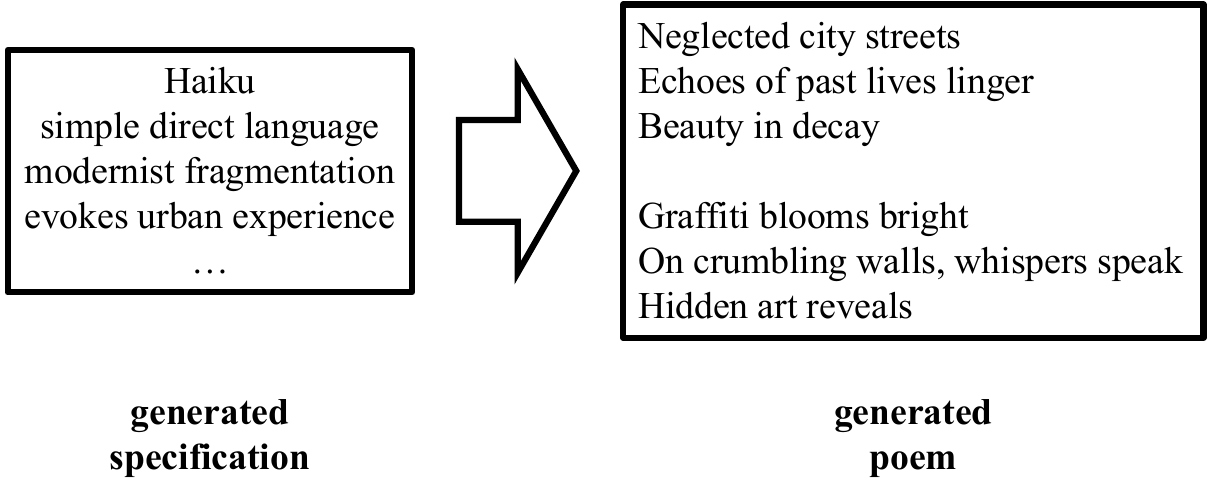}
\caption{Chain-of-Specification prompting}
\label{fig:illustration}
\end{figure}

The key challenge is how to design techniques capable of diverse generation according to a user-defined feature mapping. Inspired by chain-of-thought prompting~\cite{cot1}, we propose a two-step prompting strategy, which we call \emph{single specification (SS) prompting} (summarized in Figure~\ref{fig:illustration}). First, we use the LLM to generate an example of structure $s\in\mathcal{S}$, which we call a \emph{specification}. Second, we prompt the LLM to generate text $x\in\mathcal{X}$ that satisfies the specification $s$ (i.e., $\phi(x)=s$). This strategy isolates the source of diversity to the first step---i.e., as long as the LLM generates a diverse range of specifications $s$, then the generated text $x$ will be diverse (assuming $\phi(x)=s$ always holds). Intuitively, generating a diverse range of specifications is easier than directly generating text $x$ with diverse structure.

This strategy can also be chained, where we first generate a high-level specification, then generate a low-level specification that satisfies it, and finally generate text that satisfies the low-level specification; more levels can also be included. For example, in the poetry domain, the prompts might ask for different kinds of structure, such as style, theme, imagery, etc. This process resembles chain-of-thought prompting~\cite{cot1} since it asks the LLM to derive the final generated text using multiple steps, which we refer to as \emph{chain-of-specification (CoS) prompting}.

We evaluate our approach on domains including poetry generation, code generation, and generating coding challenge problems. Our results demonstrate that our approach is significantly more effective at improving structural diversity compared to existing approaches for diverse generation; one exception is for models that are not instruct tuned, since our approach relies on instruction following to work well. Finally, we also provide evidence that structural diversity captures qualitatively different aspects of diversity compared with existing metrics such as $n$-gram and BERT embedding diversity, demonstrating the value of structural diversity.

\para{Example.}
We've provided an example of three poems generated using each our CoS algorithm and a standard random sampling strategy in Appendix~\ref{sec:examples}. In the examples generated by random sampling, even though the words are different from one poem to another, the content and structure appear very consistent across all three samples. In constrast, the poems sampled using SoC exhibit significantly different structure and content. These kinds of differences occur in all of our domains.

\para{Contributions.}
We propose a novel framework for studying structural diversity in text generation, where diversity is defined as the entropy of a user-defined mapping into a feature space. In addition, we propose chain-of-specification (CoS) prompting, an effective algorithm for improving structural diversity. Our experiments demonstrate that our approach can significantly improve structural diversity compared to several baselines.

\para{Related work.}
Several recent studies have investigated methods for quantifying and improving the diversity of text generated by LLMs. For example, \citet{diversity1} proposes a set of metrics for evaluating the diversity of generated text, including self-BLEU, distinct $n$-grams, and entropy. They also introduced a new decoding method called nucleus sampling, which aims to improve diversity by sampling from the top-p portion of the probability distribution at each step. This approach works with any model but is white box, whereas our algorithm is black box.

\citet{diversity2} studies the trade-off between diversity and quality in text generation, using a combination of automatic metrics and human evaluations. They found that increasing the diversity of generated text often comes at the cost of reduced coherence and relevance. To address this issue, they proposed a new approach called DiversityGAN, which uses a generative adversarial network to generate diverse and high-quality text. This approach requires modifying the training process. To the best of our knowledge, the only black-box diversity improvement algorithm was developed and discussed in \cite{ippolito}, who suggested oversampling, clustering the samples into much fewer clusters using an approach such as K-Means, and then only taking the centroid from each cluster - we compare to this baseline in our experiments.

Finally, \citet{diversity3} investigates the diversity of text generated by GPT-4~\cite{chatgpt} using metrics such as $n$-gram diversity, part-of-speech diversity, and semantic diversity, to compare the diversity of GPT-4 generated text to that of human-written text. They find that while GPT-4 generates text with high ``local'' (i.e., $n$-gram) diversity, it tends to exhibit lower ``global'' (i.e., semantic) diversity compared to human-written text. However, they do not study how to bridge this gap.

More broadly, there has been work studying diversity for reasoning~\citep{naik2023diversity,zhang2024improving}, and improving diversity of recommender systems~\cite{carraro2024enhancing}.

\section{Chain-of-Specification Prompting}\label{sec:methodology}

\para{Problem formulation.}
We assume given a mapping $\phi:\mathcal{X}\to\mathcal{S}$, where $\mathcal{X}=\Sigma^*$ is the space of possible generated text, and $\mathcal{S}=\{0,1\}^d$ is a space of structures. We assume $\mathcal{S}$ is binary and define metrics accordingly, but more general spaces can be used if the metrics are correspondingly modified. For example, in the poetry domain, a latent variable could be the sentiment of the poem (idyllic, melancholic, etc.), the rhyme scheme (regular, irregular, etc.), or the meter (iambic pentameter, free verse, etc.), encoded as one-hot variables.

Our goal is to generate outputs $x\in\mathcal{X}$ with diverse structure $\phi(x)$. We focus on the setting where the large language model (LLM) $p$ is given a fixed prompt, so $p$ can be thought of as a probability distribution over $\mathcal{X}$, and we want to sample a set of diverse generations $x_1,...,x_k\sim p$. To measure diversity, we assume given a 
diversity metric $F:\mathcal{S}^k\to\mathbb{R}_{\ge0}$ (see Section~\ref{sec:experiments} for our choices of $F$); then, we define diversity to be
\begin{align*}
D=\mathbb{E}_{x_1,...,x_k\sim p}[F(\phi(x_1),...,\phi(x_k))].
\end{align*}

\para{Specification prompting.} The idea in specification prompting is to first prompt the LLM to generate a random specification $s\in\mathcal{S}$, and then prompt the LLM to generate text $x\in\mathcal{X}$ such that $\phi(x)=s$. Intuitively, it is easier for the LLM to generate random specifications (which have relatively simple structure) than to generate random text directly. We denote the LLM prompted to generate specifications as $q(\cdot)$ (which is a distribution over $\mathcal{S}$), and the LLM prompted to generate text for a given specification $s\in\mathcal{S}$ as $p(\cdot\mid s)$ (which is a distribution over $\mathcal{X}$). Then, to generate $k$ diverse samples, we first sample $s_1,...,s_k\sim q$, and then $x_i\sim p(\cdot\mid s_i)$ for each $i\in\{1,...,k\}$. We provide examples of prompts in Appendix~\ref{subsec:prompts}.

\para{Chain-of-specification prompting.}
We can straightforwardly extend specification prompting by first generating high-level specifications, then generating low-level specifications, and then generating the text. In this case, we assume the user provides mappings $\phi_j:\mathcal{S}_j\to\mathcal{S}_{j-1}$ for $j\in\{1,...,m\}$, where $\mathcal{S}_m$ is the highest level specification, $\mathcal{S}_1$ is the lowest level specification, and $\mathcal{S}_0=\mathcal{X}$. Then, for $j\in\{0,1,...,m\}$, we use $s'\sim q_j(\cdot\mid s)$ to denote sampling specification $s'\in\mathcal{S}_j$ conditioned on specification $s\in\mathcal{S}_{j+1}$ (for $j=m$, $s$ is empty).

Now, we first draw samples $s_{m,i}\sim q_m$ (for $i\in\{1,...,k\}$); then, for each $j=m-1$ to $j=0$, we draw samples $s_{j,i}\sim q_j(\cdot\mid s_{j+1,i})$. Finally, letting $x_i=s_{0,i}$ we return samples $x_1,...,x_k$.

This approach is particularly effective for domains where the desired output can be naturally decomposed into a hierarchy of specifications---e.g., for poetry generation, the high-level specifications could include poetry styles and themes, whereas the mid-level and low-level specifications could include more specific attributes such as emotional tone, imagery, or rhyme schemes.

\section{Experiments}
\label{sec:experiments}

\subsection{Experimental Setup}

\para{Datasets.}
We conducted experiments on three domains: poetry, coding challenge problem descriptions, and code solutions. The real human poetry dataset was taken from the Poetry Foundation's collection of curated material \cite{divy2021poetry}. The coding challenge problem descriptions were non-overlapping problem descriptions from Project CodeNet \cite{puri2021codenet}. The code solutions were also taken from CodeNet, with the requirement that only one solution be sampled per individual coding challenge.

In the poetry domain, features include spacing, rhyme, and meter; in the coding challenge problems domain, features include whether the problem uses matrix manipulation or whether it specifies memory constraints; and in the code solutions domain, features include whether the program uses recursion or whether it has input validation. We constructed 300 features for the poetry domain, 90 for the educational coding challenge domain, and 185 for the Python code domain.

\begin{figure*}
\centering
\includegraphics[width=0.9\textwidth]{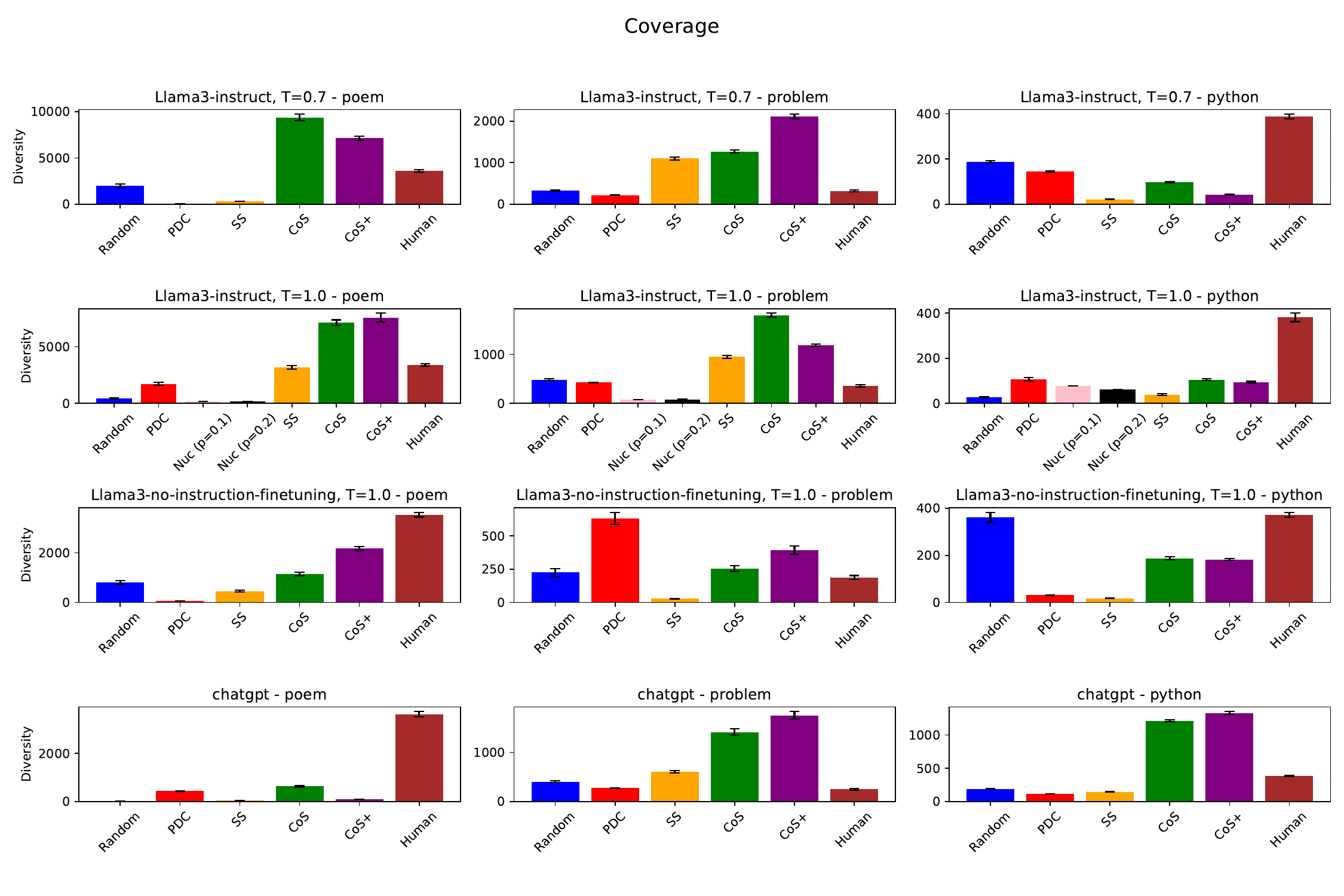}
\caption{Results of Diversity Metrics for Poetry, Code, and Coding Problem Domains, respectively.  Higher = Better.}
\label{fig:results}
\end{figure*}

\para{Approaches.}
We compare the diversity of our approach to the following:
\begin{itemize}
\item \textbf{Random sampling:} Sample (with positive temperature) from the LLM based on a default prompt that describes the target domain.
\item \textbf{Human:} Human-written text in the domain.
\item \textbf{Post-decoding clustering (PDC):} A method proposed in \cite{ippolito} to improve diversity by generating a magnitude more outputs than needed and then taking the ones corresponding to the centroids of $K$-means clustering. We use an initial sample size of 300 and $K=100$.
\item \textbf{Nucleus (top P) sampling:} A method proposed by 
\citep{diversity1} that samples text from the most likely tokens that collectively have probability at least $p$; this strategy allows for diversity while truncating the less reliable tail of the distribution.
\end{itemize}

\para{Metrics.}
To measure structural diversity, we use metrics that capture the coverage of the feature space $\mathcal{S}$. Given a set of generations $x_1,...,x_k$, let $S = \{s_1, ..., s_k\}\subseteq\mathcal{S}$ be the set of corresponding structures, where $s_i=\phi(x_i)$. We assume that for each $j \in \{1, ..., d\}$, feature $j$ is present in at most half of the samples---i.e., $s_{ij} = 1$ for at most half of the $s_i$. This assumption holds in our experiments since structural features tend to be difficult to satisfy. Then, we define \emph{coverage} to be
\begin{align*}
C_n(S) = \frac{\sum_{c \in \mathcal{C}_n} \log(|\{s \in S : c \subseteq s\}| + 1)}{\log(|S| + 1)},
\end{align*}
where $\mathcal{C}_n$ is the set of all possible combinations of $n$ features for some given hyperparameter $n$; we use $n=3$ in our experiments included below. In other words, it is the sum of logarithmically weighted feature combination counts, normalized by the maximum possible weighted count. Intuitively, high coverage indicates that the samples capture many different feature combinations. By our assumption that $j$ is present in at most $1/2$ of samples, most feature combinations are rare, meaning structural diversity is required to achieve high coverage.

We provide results for variations of our coverage metric, as well as standard $n$-gram and BERT embedding diversity metrics, in Appendix~\ref{sec:additionalexperiments}. 

\para{Language models.} We evaluate four LLMs: ChatGPT-3.5-turbo at a temperature of 1.0, Llama3-70B-Instruct~\citep{touvron2023llama} at a temperature of 1.0, Llama3-70B-Instruct at a temperature of 0.7, and vanilla Llama3-70B (i.e., not instruct tuned) at a temperature of 1.0. We provide prompts in Appendix~\ref{subsec:prompts}.

\para{Samples.} For each LLM, we took $k=300$ samples in $S$, and report mean and standard errors over $300$ samples when bootstrapping with sub-sampling 50 samples per iteration. Similarly, the human datasets were evaluated using 300 random samples.

\subsection{Experimental Results}

Figure~\ref{fig:results} shows results for each domain (column), LLM configuration (row), and approach (bar).

\para{Comparison to baselines.} CoS almost always outperforms all our baselines. The main exception is Llama without instruction tuning, which is expected sicne our approach relies on instruction following to be effective. In many cases, SS also performs well in both the poetry and problem domains, though it performs worse in the Python domain, likely because the specifications are more difficult to follow (e.g., using loops vs. recursion). In some cases, incorporating PDC and CoS (i.e., CoS+) produces a small additional benefit, though it can also sometimes reduce performance, indicating that its effectiveness is domain specific.

\para{Comparison to human.} For the poetry and problem domains, CoS matches or even slightly exceeds the human dataset in terms of diversity, highlighting the effectiveness of our approach.

\para{Comparison on existing metrics.} We show results on $n$-gram and BERT embedding diversity in Appendix~\ref{sec:additionalexperiments}. The diversity of our approach is still high according to these metrics, though the baselines often perform similarly well; for instance, random sampling is competitive with our approach in many instances. One important observation is that the human dataset is sometimes significantly less diverse than the LLMs according to these metrics; in addition, for BERT embedding diversity, GPT-4 tends to be less diverse despite being a stronger model. These trends suggest that these diversity metrics represent qualitatively different forms of diversity compared to structural diversity. The specific kind of diversity may be domain dependent, but structural diversity has the key advantage that the user can tailored it to their domain.

\para{Comparison across models.}
In general, instruction tuning tends to improve diversity. The relationship between temperature and diversity is more complicated; generally, temperature increases diversity at the token level, but it can make it harder to satisfy structures leading to lower structural diversity. Finally, while GPT-4 generally exhibits more diversity, especially when using CoS prompting, except in the poetry domain.

\section{Conclusion}\label{sec:conclusion}

We have proposed a novel framework for improving the structural diversity of black box LLMs where the user provides features encoding desired structural diversity properties, and then we use chain-of-specification prompting to automatically generate diverse outputs. Our experiments demonstrate that our framework is effective at improving structural diversity, 
Wwe also find evidence that structural diversity is qualitatively different from more traditional metrics such as $n$-gram diversity and diversity of BERT embeddings.

\section{Limitations}

One limitation of our approach is that it requires the user to design the feature mapping from text to structures. While this mapping gives the user significant control over the kind of diversity they care about, it requires additional effort for each new domain where our technique is applied. For many domains, it may be possible to automate parts of this effort, for instance, by asking a strong model such as GPT-4 to identify reasonable structural features in new domains. In addition, generating chains of specifications requires sampling significantly more tokens compared to random sampling. The benefit is that our work can be generally applicable to the black box setting. In the white box LLM setting, finetuning techniques might enhance diversity without the need to sample additional tokens.

\bibliography{references}

\clearpage
\onecolumn
\appendix

\section{Examples}
\label{sec:examples}

The following are examples generated by random sampling:

\begin{center}
\includegraphics[width=\textwidth]{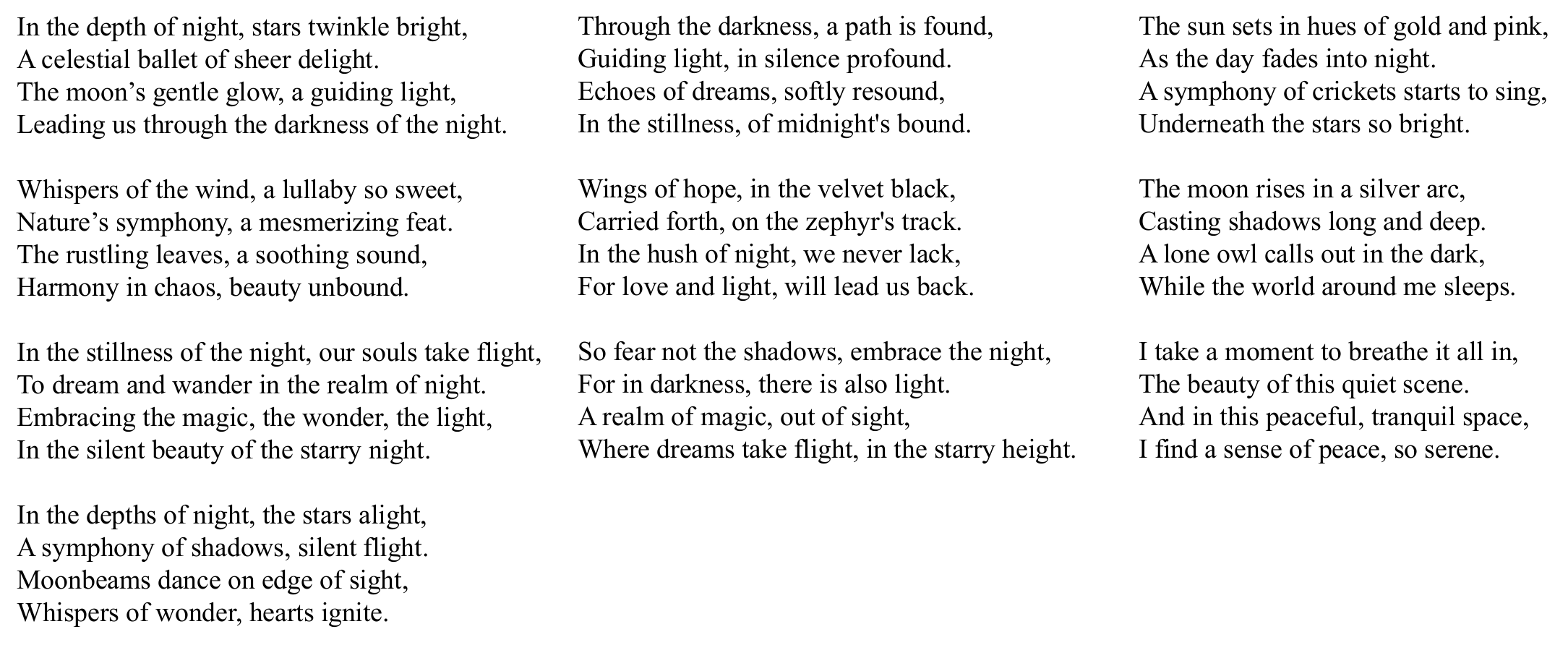}
\end{center}

\noindent The following are examples generated by CoS:

\begin{center}
\includegraphics[width=\textwidth]{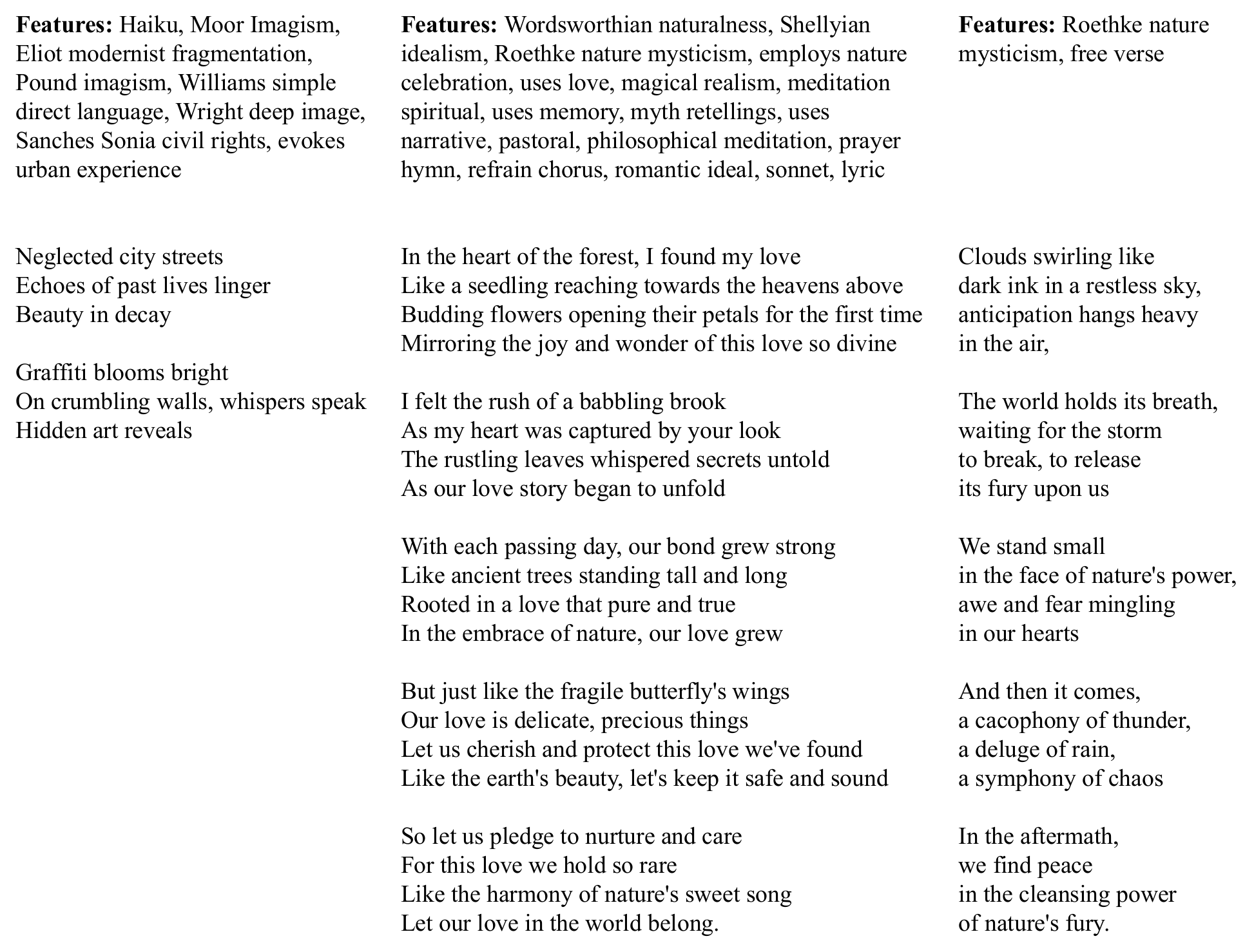}
\end{center}

\section{Additional Experiments}
\label{sec:additionalexperiments}

\subsection{Additional Metrics}

We consider the following metrics for measuring diversity.

\para{Coverage.} The sum of logarithmically weighted feature combination counts, normalized by the maximum possible weighted count:
$$C_n(S) = \frac{\sum_{c \in \mathcal{C}_n} \log(|\{s \in S : c \subseteq s\}| + 1)}{\log(|S| + 1)}$$
where $\mathcal{C}_n$ is the set of all possible combinations of $n$ features. This metric measures the extent to which the samples cover the range of possible improbable structural properties, ensuring that the generated text exhibits a diverse set of rare features.

\para{Weighted surprisal.} The average surprisal of each feature combination, weighted by its probability:
$$WS_n(S) = -\sum_{c \in \mathcal{C}_n} \frac{|\{s \in S : c \subseteq s\}|}{|S|} \log\left(\frac{|\{s \in S : c \subseteq s\}|}{|S|}\right),$$
This metric quantifies the unexpectedness of the observed combinations of improbable structural properties, giving higher weight to rare feature combinations. It ensures that the generated text contains surprising and informative structures.

\para{Boosted Jaccard diversity.} The sum of Jaccard distances between all pairs of feature combinations, weighted by their sizes and normalized by the total number of samples:
$$D_{BJ,n}(S) = \sum_{\substack{c_i, c_j \in \mathcal{C}_n \\ i \neq j}} \frac{|c_i|\cdot|c_j|}{|S|^2} \left(1 - \frac{|\{s \in S : c_i \subseteq s \wedge c_j \subseteq s\}|}{|\{s \in S : c_i \subseteq s \vee c_j \subseteq s\}|}\right).$$
This metric measures the dissimilarity between pairs of feature combinations, giving higher weight to larger combinations. It ensures that the generated samples have distinct sets of improbable structural properties, promoting diversity in the text's rare features.

\para{Dice Diversity.} The average Dice distance between all pairs of feature combinations:
$$D_{Dice,n}(S) = \frac{2}{|\mathcal{C}_n|(|\mathcal{C}_n| - 1)} \sum_{\substack{c_i, c_j \in \mathcal{C}_n \\ i \neq j}} \left(1 - \frac{2\cdot|\{s \in S : c_i \subseteq s \wedge c_j \subseteq s\}|}{|\{s \in S : c_i \subseteq s\}| + |\{s \in S : c_j \subseteq s\}|}\right).$$
This metric quantifies the dissimilarity between pairs of feature combinations using the Dice coefficient, which emphasizes the presence of rare features in both combinations. It ensures that the generated text has a diverse set of improbable structural properties that are not frequently shared between samples.

\para{One-way inclusion diversity.} The average one-way inclusion coefficient between all pairs of feature combinations:
$$D_{OWI,n}(S) = \frac{2}{|\mathcal{C}_n|(|\mathcal{C}_n| - 1)} \sum_{\substack{c_i, c_j \in \mathcal{C}_n \\ i \neq j}} \left(1 - \frac{|\{s \in S : c_i \subseteq s \wedge c_j \subseteq s\}|}{\min(|\{s \in S : c_i \subseteq s\}|, |\{s \in S : c_j \subseteq s\}|)}\right)$$
This metric measures the dissimilarity between pairs of feature combinations using the one-way inclusion coefficient, which quantifies the proportion of rare features in one combination that are not present in the other. It ensures that the generated text has a diverse set of improbable structural properties that are not subsumed by other samples.

\para{Weighted overlap diversity:} The average overlap coefficient between all pairs of feature combinations, weighted by their sizes and normalized by the total number of combinations:
$$D_{WO,n}(S) = \frac{2}{|\mathcal{C}_n|(|\mathcal{C}_n| - 1)} \sum_{\substack{c_i, c_j \in \mathcal{C}_n \\ i \neq j}} \frac{|\{s \in S : c_i \subseteq s \wedge c_j \subseteq s\}|}{\min(|\{s \in S : c_i \subseteq s\}|, |\{s \in S : c_j \subseteq s\}|)}$$
This metric quantifies the similarity between pairs of feature combinations using the overlap coefficient, which measures the proportion of shared rare features. By subtracting this metric from 1, we obtain a diversity measure that ensures the generated text has a diverse set of improbable structural properties with minimal overlap between samples.

\para{$n$-gram Diversity.}
The $n$-gram diversity of a set of generated texts is the Shannon entropy of the distribution of $n$-grams across those texts \cite{evaleval}. For instance, for $n=4$, it is
\[
D = -\sum_{i=1}^{M} p_i \log p_i
\]
where $p_i$ is the probability of occurrence of the $i$th 4-gram among all 4-grams, and $M$ is the total number of unique 4-grams in the documents. This metric quantifies the unpredictability of the text based on the variety of its 4-gram constructs, with higher values indicating more diverse generations.

\para{BERT Embedding Diversity.}
We measure the diversity of text documents based on the variability in their BERT embeddings, as per \cite{evaleval}. This approach utilizes the pre-trained BERT model to convert textual data into high-dimensional vectors, where each vector represents the semantic content of a text. In particular, the diversity is the pairwise cosine distances between the BERT embeddings of all generated texts. First, each text is transformed into an embedding by averaging the output vectors (i.e., BERT's last hidden layer) of all tokens in it. Then, we compute the cosine distances between every pair of embeddings to form a distance matrix. Finally, the BERT diversity is the mean of all of these pairwise distances. 

\subsection{Additional Results}
\label{subset:additional_results}

\begin{figure*}[h!]
\centering
\includegraphics[width=0.9\textwidth]{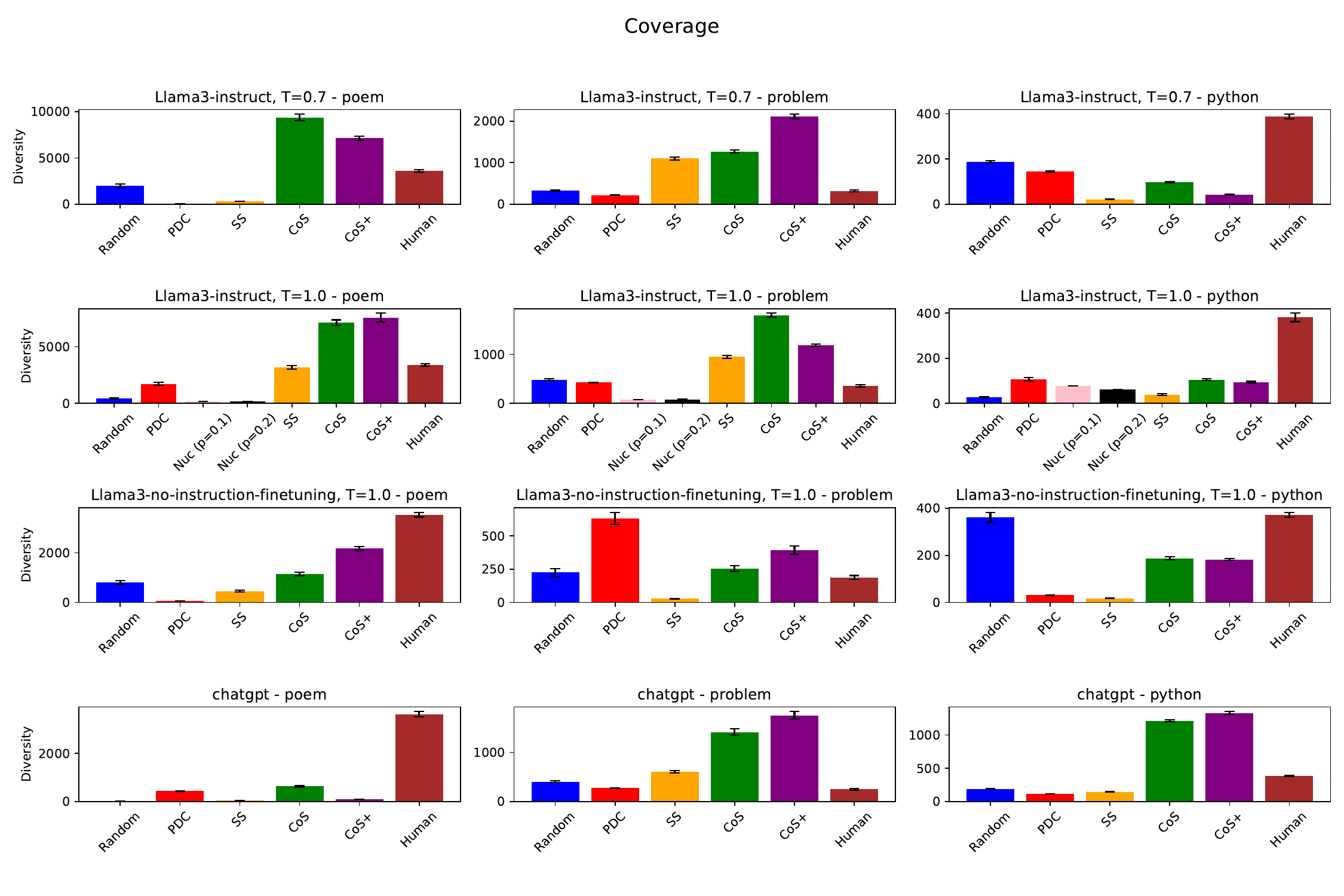}
\caption{Results of coverage diversity for Poetry, Code, and Coding Problem Domains, respectively.  Higher = Better.}
\label{fig:results_coverage}
\end{figure*}

\begin{figure*}
\centering
\includegraphics[width=0.9\textwidth]{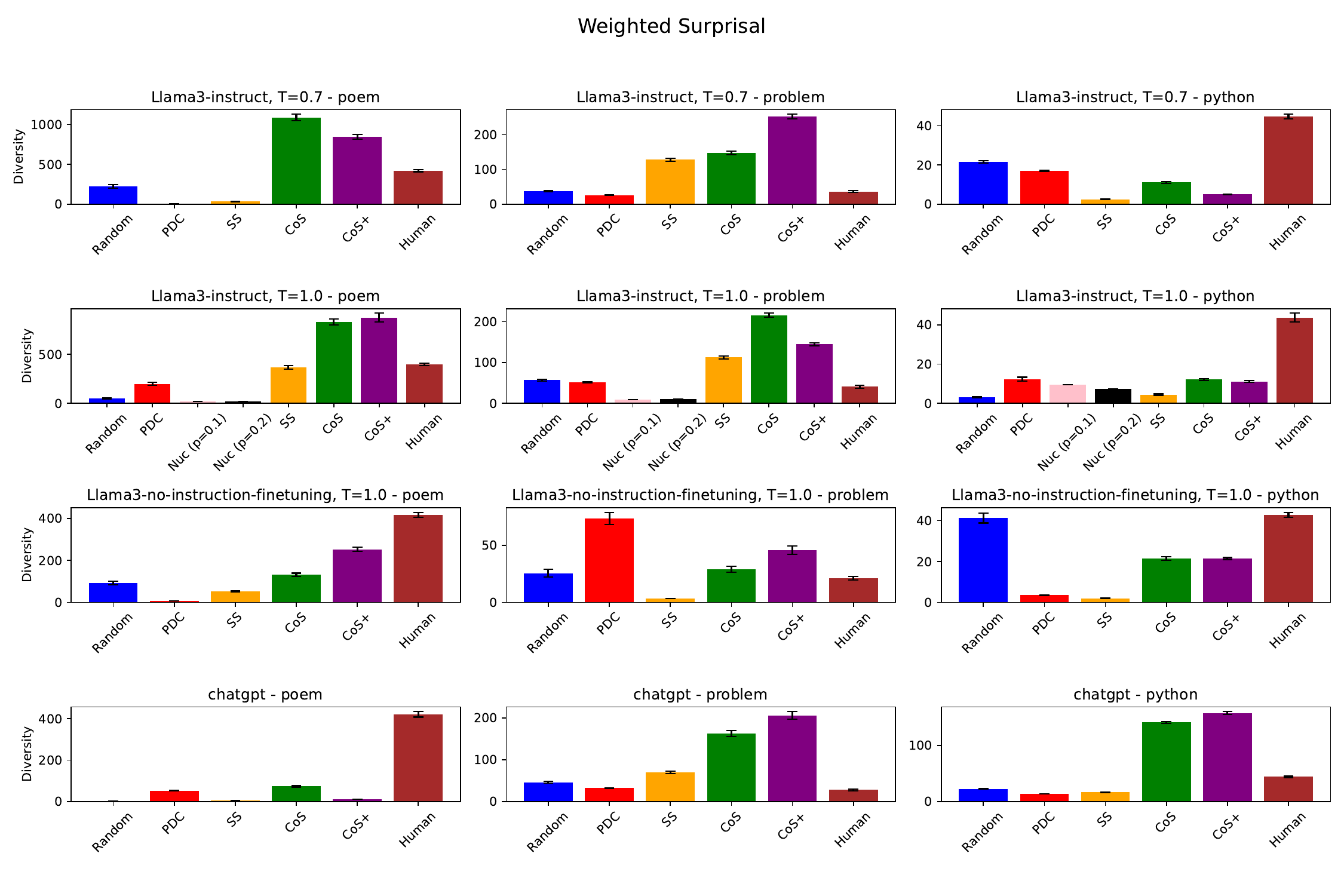}
\caption{Results of weighted surprisal diversity for Poetry, Code, and Coding Problem Domains, respectively.  Higher = Better.}
\label{fig:results_surprisal}
\end{figure*}

\begin{figure*}
\centering
\includegraphics[width=0.9\textwidth]{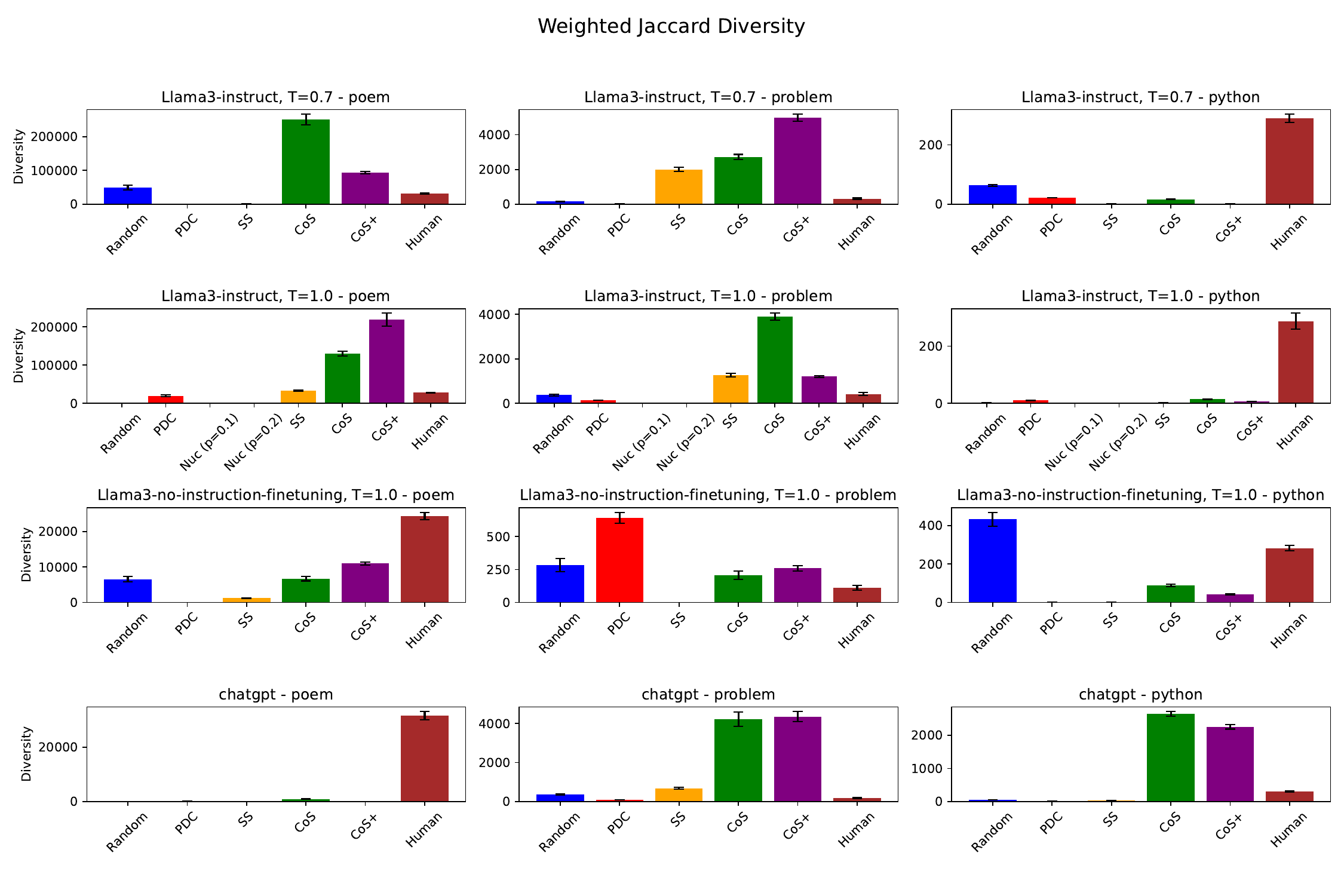}
\caption{Results of boosted Jaccard diversity for Poetry, Code, and Coding Problem Domains, respectively.  Higher = Better.}
\label{fig:results_Jaccard}
\end{figure*}

\begin{figure*}
\centering
\includegraphics[width=0.9\textwidth]{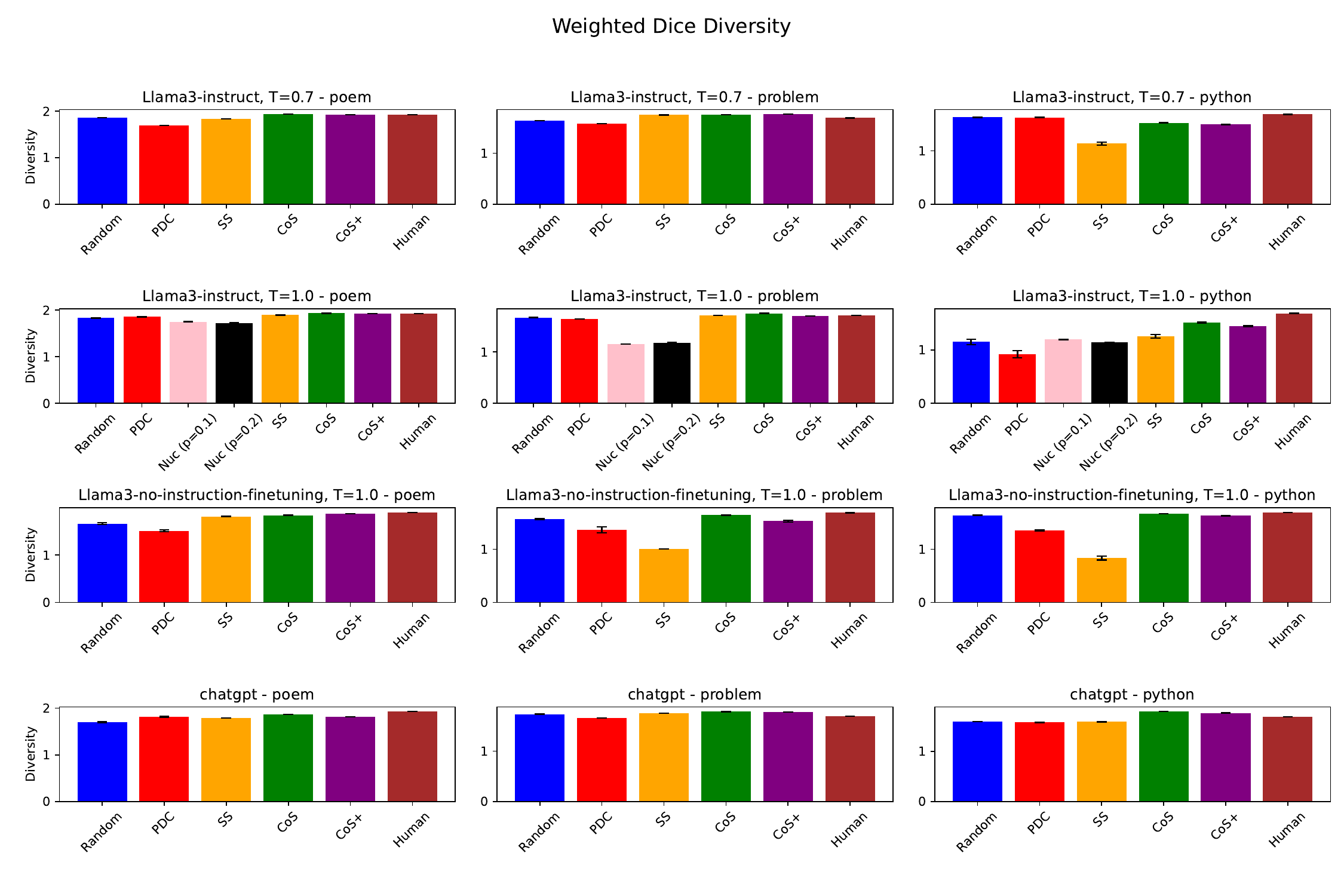}
\caption{Results of Dice diversity for Poetry, Code, and Coding Problem Domains, respectively.  Higher = Better.}
\label{fig:results_Dice}
\end{figure*}

\begin{figure*}
\centering
\includegraphics[width=0.9\textwidth]{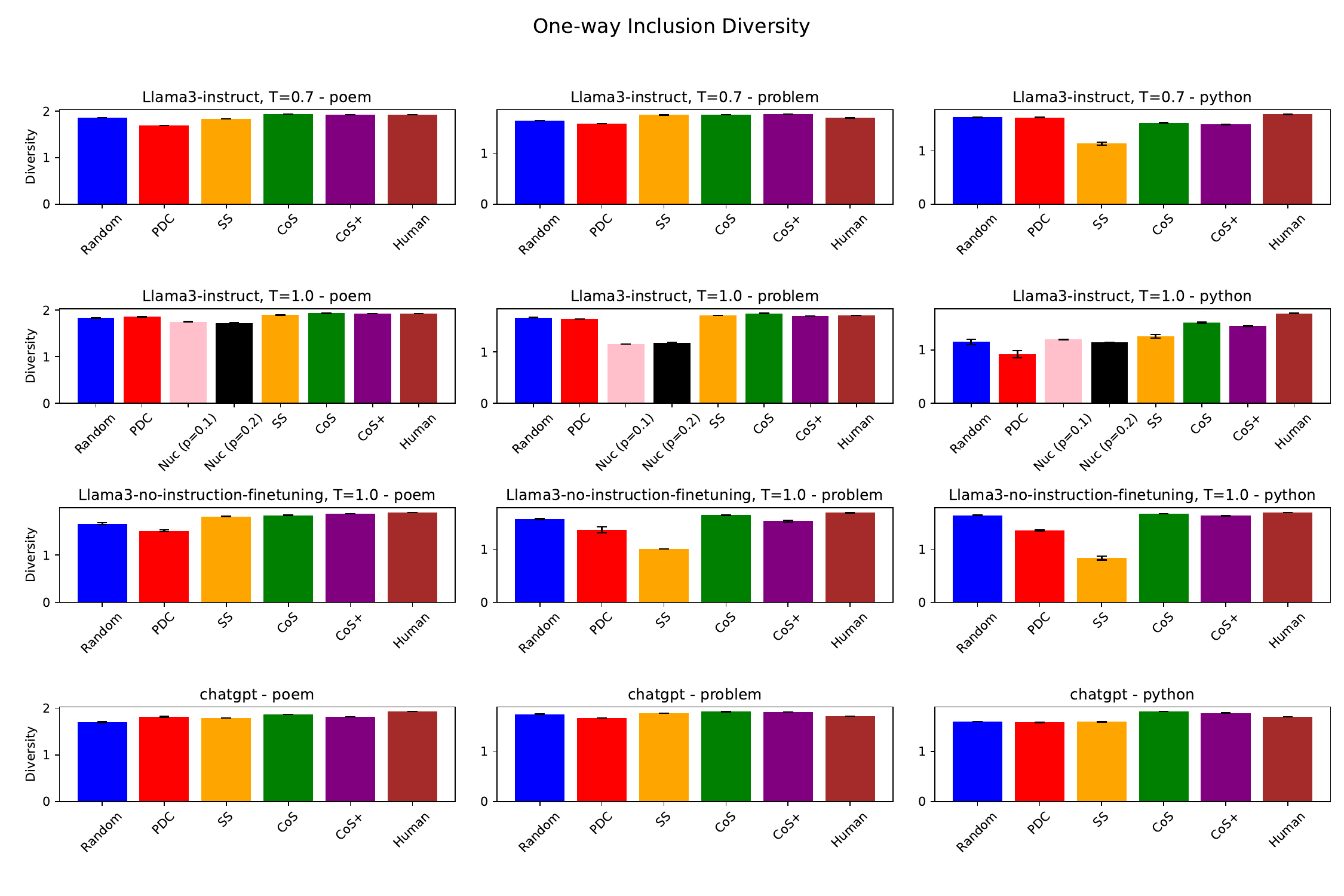}
\caption{Results of one way inclusion diversity for Poetry, Code, and Coding Problem Domains, respectively.  Higher = Better.}
\label{fig:results_inclusion}
\end{figure*}

\begin{figure*}
\centering
\includegraphics[width=0.9\textwidth]{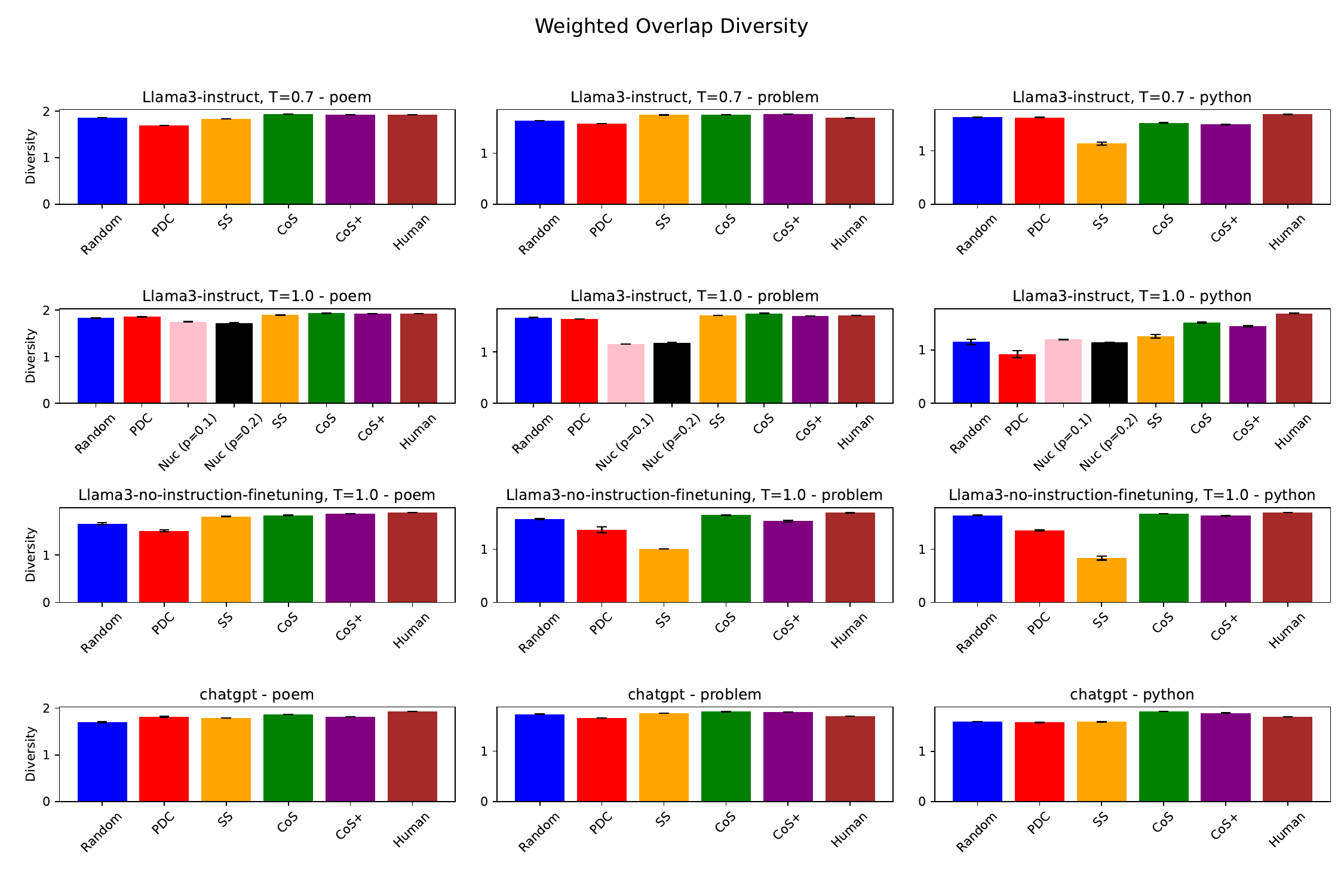}
\caption{Results of weighted overlap diversity for Poetry, Code, and Coding Problem Domains, respectively.  Higher = Better.}
\label{fig:results_overlap}
\end{figure*}

\begin{figure*}
\centering
\includegraphics[width=0.9\textwidth]{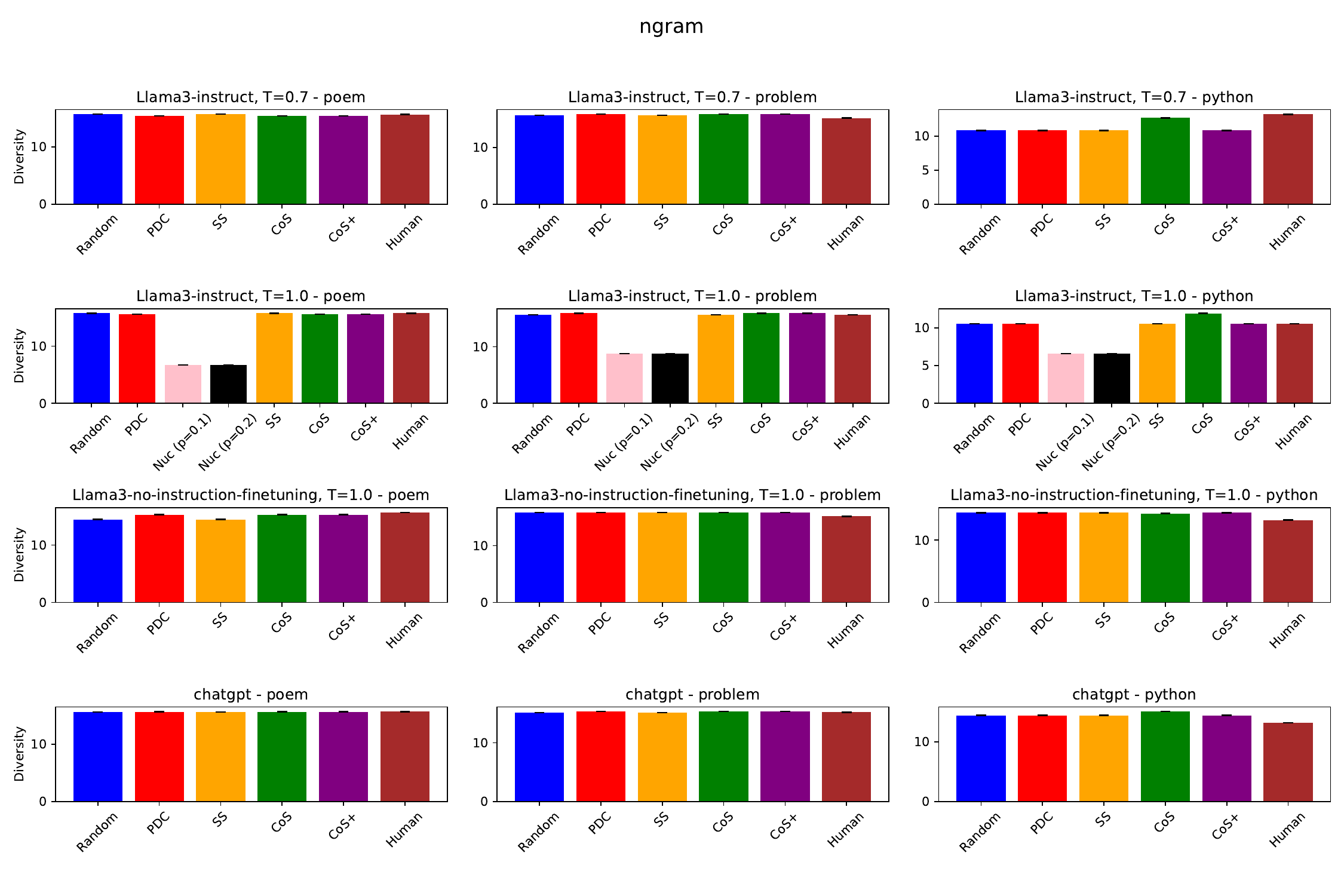}
\caption{Results of $n$-gram diversity for Poetry, Code, and Coding Problem Domains, respectively.  Higher = Better.}
\label{fig:results_ngram}
\end{figure*}

\begin{figure*}
\centering
\includegraphics[width=0.9\textwidth]{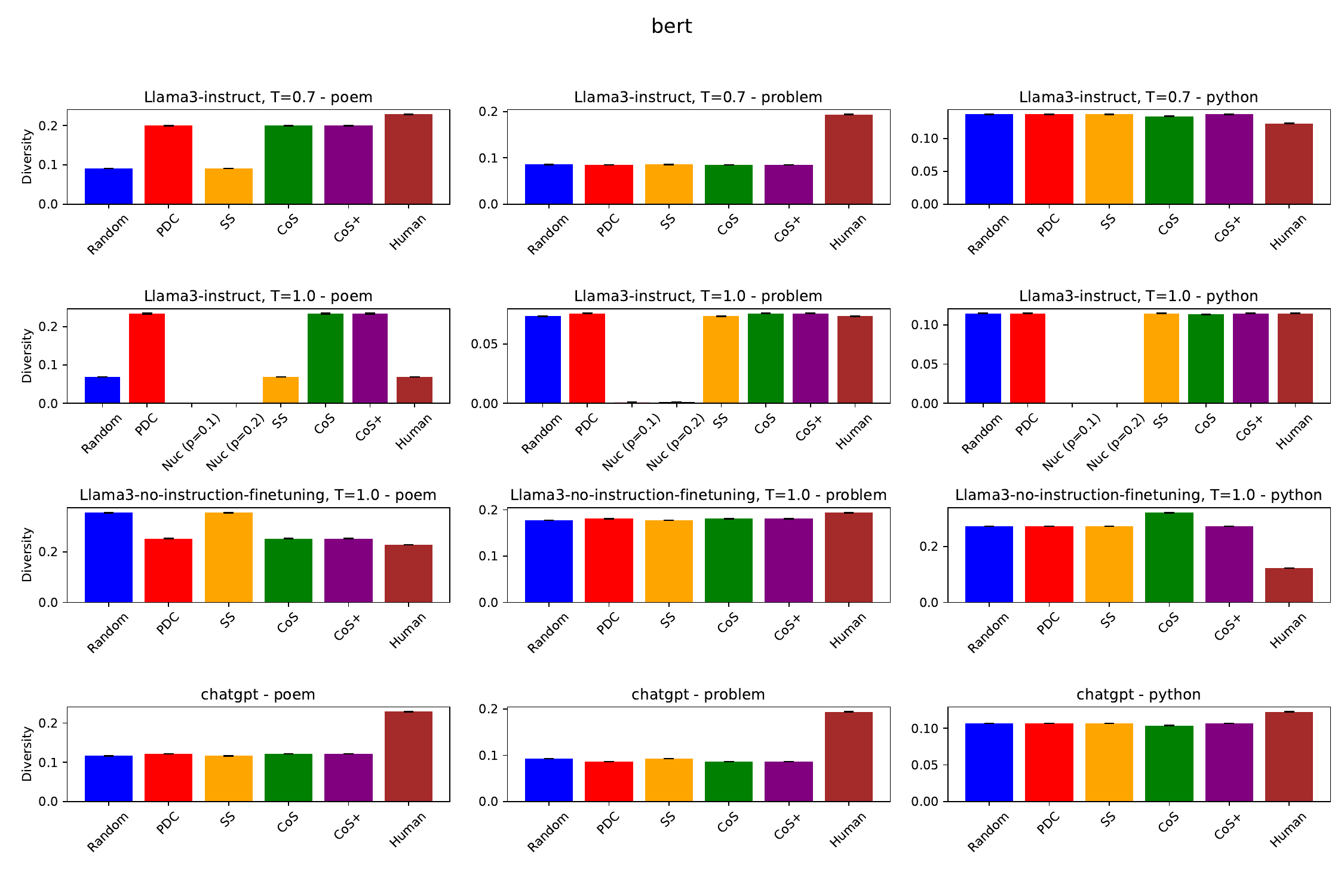}
\caption{Results of \textbf{BERT} diversity for Poetry, Code, and Coding Problem Domains, respectively.  Higher = Better.}
\label{fig:results_bert}
\end{figure*}

We show results for each of the additional metrics in Figures~\ref{fig:results_coverage}, \ref{fig:results_surprisal}, \ref{fig:results_Jaccard}, \ref{fig:results_Dice}, \ref{fig:results_inclusion}, \ref{fig:results_overlap}, \ref{fig:results_ngram}, \&~\ref{fig:results_bert}.

\section{Prompts}
\label{subsec:prompts}

We show the prompts used for each of our three domains in Figures~\ref{fig:poem_prompting}, \ref{fig:problem_prompting}, \&~\ref{fig:python_prompting}.

\begin{figure}[h]
\textcolor{blue}{LLM Prompt:} Come up with an interesting and original prompt for a poem.  Return the prompt only; no prefix or commentary. \\

\textcolor{blue}{LLM Style:} Suggest a poetic style that fits and does not at all contradict, but rather complements and is integrated into this prompt {prompt}.  Describe what the poem does or doesn't do relative to 'traditional' poetry, and how this relates to the prompt.  Don't limit yourself in any way - consider styles from Shakespeare to contemporary eco-concrete-poetry and modernist whitespace, concrete, and fragmentation poetry, and from all over the world. Consider influences as varied as John Dunne and Charles Bernstein or ee. Cummings. Or, interpolate between existing styles, or invent a new poetic style and describe it. Note that you should prioritize finding a way to satisfy the prompt {prompt} over all other constraints, and attempt to strictly enhance that prompt. Just return the description of the poetic style - no commentary. \\

\textcolor{blue}{LLM Theme:} Given the poetic style '{style}' (which you may or may not be familiar with) and this prompt: {prompt}, suggest a suitable theme, and elaborate how it will be presented in relation to the form in detail. Note that you should prioritize finding a way to satisfy the prompt {prompt} over all other constraint, and attempt to strictly enhance that prompt.  Return only the theme, no prefix or commentary. \\

\textcolor{blue}{LLM Mood:} Given the theme '{theme}' and the style {style} and the prompt {prompt}, suggest a congruent (yet not necessarily the most obvious) emotional arc to the poem.  Note that you should prioritize finding a way to satisfy the prompt {prompt} over all other constraint. Return only the emotional arc and how it fits with the prompt as a paragraph, no prefix or commentary. \\

\textcolor{blue}{LLM Imagery:} Given the theme '{theme}' and the style '{style}' and the emotional arc '{mood}' and the prompt {prompt}, provide one possible type of imagery to include in the poem. Note that you should prioritize finding a way to satisfy the prompt {prompt} over all other constraint.  Return only the imagery, no prefix or commentary. \\

\textcolor{blue}{LLM Poem:} Compose a poem with this prompt: {prompt} in the form of {form}, exploring the theme '{theme}', conveying a '{mood}' emotional arc, and incorporating this imagery: '{imagery}'. Note that you should prioritize finding a way to satisfy the prompt {prompt} over all other constraint.  Return only the poem, no prefix or commentary.
\caption{Prompts used for CoS+PDC sampling of poems.}
\label{fig:poem_prompting}
\end{figure}

\begin{figure}[h]
\textcolor{blue}{LLM Types:} Write an example input-output type pair for a python programming challenge.  Return only the input type and output type; no prefix or commentary. \\

\textcolor{blue}{LLM Goal:} Write an educational goal for a python programming challenge.  You are constrained in one way: The input-output types must be {types}.Some examples might be teaching a particular lesson about recursion, or teaching about the importance of programming efficiently, but any educational goal within computer science could work. Return the educational goal description as a paragraph only; no prefix or commentary. \\

\textcolor{blue}{LLM CoS Program:} Write a python program which satisfies the following educational goal: {goal} and has the following input-output-types: {types}. Return the python program only; no prefix or commentary. \\

\textcolor{blue}{LLM SS Program:} Write a python program which satisfies the following input-output type: {types}. Return the python code alone; no prefix or commentary. \\

\textcolor{blue}{LLM Random Program:} Write a 100-line python program.  Return the code only; no prefix or commentary.
\caption{Prompts used for CoS+PDC/SS/Random sampling of Python programs.}
\label{fig:python_prompting}
\end{figure}

\begin{figure}[h]
\textcolor{blue}{LLM Goal 1:} Write an educational goal for a programming challenge.  Some examples might be teaching a particular lesson about recursion, or teaching about the importance of programming efficiently, but any educational goal within computer science could work. Return the educational goal description as a paragraph only; no prefix or commentary. \\

\textcolor{blue}{LLM Goal 2:} Conditioned on wanting to teach about {Goal 1}. Write a secondary educational goal you might have for a coding challenge.  Return a paragraph-long description of what you're trying to achieve pedagogically, in addition to: {Goal 1}. Return it as a paragraph without prefix or commentary. \\

\textcolor{blue}{LLM CoS problem description:} Write an example coding challenge problem which could work for a programming teacher who wants to teach primarily about the following: {Goal 1} and secondarily about the following: {Goal 2}. Make it as descriptive as possible, including a description of the problem, example input-output, and any additional information that may be needed.  Note that it should be programming-language agnostic. \\

\textcolor{blue}{LLM SS Problem Description:} Write an example coding challenge problem which could work for a programming teacher who wants to teach about the following: {Goal 1}. Make it as descriptive as possible, including a description of the problem, example input-output, and any additional information that may be needed.  Note that it should be programming-language agnostic. \\

\textcolor{blue}{LLM Random Problem Description:} Write an example educational coding challenge problem.  Make it as descriptive as possible.
\caption{Prompts used for CoS+PDC/SS/Random sampling of coding challenge problem descriptions.}
\label{fig:problem_prompting}
\end{figure}

\end{document}